\documentclass{article}
\usepackage{spconf,amsmath,graphicx}
\usepackage{enumitem}
\usepackage{cite}
\usepackage{float}
\usepackage{multicol}
\usepackage{blindtext}
\usepackage{dblfloatfix}
\usepackage{tabularx}
\usepackage{amsmath}
\usepackage{lipsum}
\title{Feature Encoding in Band-limited Distributed Surveillance Systems}
\name{Alireza Rahimpour, Ali Taalimi, Hairong Qi}
\address{Department of Electrical Engineering and Computer Science\\
	University of Tennessee, Knoxville, TN, USA 37996\\
	$ \{ $arahimpo, ataalimi, hqi$ \} $@utk.edu}
\begin{document}
%\ninept
%
\maketitle
\begin{abstract}
	\vspace{0.5 cm}
Distributed surveillance systems have become popular in recent years due to security concerns. However, transmitting high dimensional data in bandwidth-limited distributed systems becomes a major challenge. In this paper, we address this issue by proposing a novel probabilistic algorithm  based on the divergence between the probability distributions of the visual features in order to reduce their dimensionality and thus save the network bandwidth in distributed wireless smart camera networks. We demonstrate the effectiveness of the proposed approach through extensive experiments on two surveillance recognition tasks. 

\end{abstract}
\section{Introduction}
\label{sec:intro}
As the concerns about public safety increase in recent years (due to some incidents such as Boston bombing, etc.), researchers have focused more on developing surveillance systems based on distributed wireless smart cameras. These cameras can cooperate, forming a wireless visual sensing network whose nodes besides visual sensing, also have processing, storage and communication capabilities.
Because the smart camera networks have become increasingly more affordable and perform better in balancing the computational power and energy efficiency, they have been employed in many surveillance tasks including distributed object recognition \cite{redondi2015cooperative}, \cite{christoudias2008unsupervised}, \cite{ferrari2004integrating}, \cite{rahimpour2016distributed}, cross view action recognition \cite{rahmani2015learning}, \cite{zheng2016cross} and person re-identification \cite{lisanti2015person}, \cite{ahmed2015improved} to name just a few. 

However, a major challenge in visual sensor networks is limitation in terms of transmission bandwidth, storage and processing power. In the traditional system design for visual sensor networks, images are acquired and compressed locally at the camera nodes, and then transmitted to the base station which performs the specific analysis tasks (e.g., video surveillance, object recognition, etc.). However, recently, a new paradigm has emerged based on analyze-then-compress, where the visual content is processed locally at the camera nodes, to extract a concise representation constituted by local visual features (e.g., SIFT, SURF, HOG). Such features are then compressed and transmitted to the base station for further analysis. Since the feature-based representation is usually more compact than the pixel-based representation, the analyze-then-compress approach is particularly attractive for those scenarios for which the bandwidth is scarce \cite{redondi2015cooperative}. 

Recently several works have been done towards the analyze-then-compress feature compression. For instance, in \cite{mitra2014toward} and \cite{naikal2010towards}, authors explore approaches for scalability in large-scale camera networks using recent advances in Compressive Sensing (CS). In \cite{naikal2011informative}, the dimension of the features is reduced using an approach based on Sparse Principal Component Analysis (SPCA). 
Furthermore, \cite{yeo2008rate} argued that reliable feature correspondence can be established in a much lower dimensional space between cameras, even if the feature vectors are linearly projected onto a random subspace.
\cite{christoudias2008unsupervised} studied a SIFT-feature selection algorithm, where the number of SIFT features that need to be transmitted to the base station is reduced by considering the joint distribution of the features among multiple camera views of a common object.
 
Another study \cite{turcot2009better}, further considered using robust structure-from-motion techniques (e.g., RANSAC) to select strong object features between two camera views, and subsequently rejecting weak features from the final stage of object recognition. Moreover, some variant works such as \cite{dong2015multi} focused on multi-view feature engineering and learning (e.g., introducing some new descriptors such as Multi-View HOG). 
However, such solutions are known to break
down easily when the camera transformation is large or when the features are extracted from low-quality images. Moreover, most of the existing approaches (e.g., \cite{turcot2009better}, \cite{christoudias2008unsupervised}) require the communication between the smart cameras for selecting the best visual features. 

The contribution of this work is three-fold. First, we propose a probabilistic algorithm  based on the divergence between the probability distributions of the visual features in order to select the most informative visual features and building a compact and physically meaningful model of the training set. Second, we introduce a scheme for calculating the low-dimensional codes for visual feature of each image, before its transmission to the base station and without any communications between the cameras. Third, we elaborate on the distributed recognition task and illustrate the performance of the proposed approach based on the experiments on two challenging and low-resolution multi-view datasets.  

The remainder of this paper is organized as follows. Section \ref{Sec: method} elaborates on the proposed method for obtaining a compact model of the feature histograms in the off-line training stage and then introduces a scheme which uses this compact representation of the training set to encode the histogram of features to low-dimensional codes. Section \ref{Sec: Exp} describes the experimental setting, followed by detailed discussion and comparison of the results. The final section concludes the paper.

\section{Methodology }\label{Sec: method}
%\subsection{Distributed Recognition System}
%\label{SubSec: dist recognition}
In our distributed recognition system, dense SIFT feature descriptors \cite{lowe1999object} are computed at a grid of overlapped patches in the image and further quantized to form a dictionary of visual words using the bag-of-words (BoW) approach \cite{lee2008libpmk}. Using the hierarchical k-means, all the feature descriptors are clustered into visual words and a term-frequency visual histogram is defined for each image. After performing the feature encoding via the proposed method, the low dimensional encoded feature histogram of the test image is sent to the base station and a nearest neighbor search (using the chi-square distance) is conducted in the training set to find the closest histogram to the testing sample. The class label of the closest histogram in the training set is then used to label the histogram of the test data.
\subsection{Compact Representation of the Features} \label{SubSec: Compact}

In recent years several studies have been carried out in the context of compact dictionary learning \cite{jiang2013label,kong2012dictionary,jiang2012submodular}, as an approach for finding a compact representation of the data. However, the lack of physical interpretation of the compact dictionary (i.e., physical meaning of each basis in the dictionary) has been a critical shortcoming of the standard dictionary learning techniques. In this paper, we address this issue by proposing a novel probabilistic approach for selecting a group of features as a compact representation of all the features in the training set. We believe that nothing is more meaningful for representing the data than the data itself. 

Assume there are $c$ classes in the training set and there are $N$ feature histograms $\boldsymbol{h}_i\in R^{m}$ in each class (i.e., $H=\{\boldsymbol{h}_1,\dots, \boldsymbol{h}_N\} \in R^{m\times N}$). When each bin of histogram is divided by the number of visual words in each cluster, the probability density function (\textit{pdf}) which represents a probability distribution is produced. Therefore, for each class we have $N$ \textit{pdf}s as: $F=\{\boldsymbol{f}_1,\dots, \boldsymbol{f}_N\} \in R^{m\times N}$.

The objective of this stage of the proposed approach is to compare all the \textit{pdf}s in each class and select a few informative ones by solving the following optimization problem: 
\begin{equation} \label{eq:first main opt eq}
\centering
\begin{gathered}
\operatornamewithlimits{min}\limits_{w_{ij}}{\sum_{j=1}^{N}\sum_{i=1}^{N}(\sum_{k=1}^{m}(( \boldsymbol{f}_i(k)-\boldsymbol{f}_j(k)) \ln (\frac{\boldsymbol{f}_i(k)}{\boldsymbol{f}_j(k)}))w_{ij}}   \\
s.t.~ \sum_{i=1}^{N} w_{ij}=1,~\forall{j}; ~~
 (\sum_{i=1}^{N}(\sum_{j=1}^{N}|w_{ij}|^q)^{p/q})^{1/p} \leq \lambda \\~~~w_{ij}\geq{0}, ~\forall{i,j}, 
%~~~\Arrowvert{W}\Arrowvert_{p,q}=k
\end{gathered}
\end{equation} 
where $\sum_{k=1}^{m}(( \boldsymbol{f}_i(k)-\boldsymbol{f}_j(k)) \ln (\frac{\boldsymbol{f}_i(k)}{\boldsymbol{f}_j(k)})$ is the symmetric form of the KL divergence \cite{kullback1951information}. This term measures the difference between all the probability distribution pairs in $F$.
     
$w_{ij}$ is defined as the probability of $\boldsymbol{f}_i$ being a representative for $\boldsymbol{f}_j$ (i.e., $w_{ij} \in [0,1]$). Therefore, we must have $\sum_{i=1}^{N} w_{ij}=1$, to assure that the probability of each $\boldsymbol{f}_j$ being represented via $F=\{\boldsymbol{f}_1,\dots, \boldsymbol{f}_N\}$ is equal to one.
Hence, the first term in Eq. \ref{eq:first main opt eq} is the cost of representing $\boldsymbol{f}_j$ via $\boldsymbol{f}_i$, which is defined as the divergence measure between them, times the probability of the occurrence of this event.  
We define $\boldsymbol{W}\in R^{N\times N}$ as the probability matrix for all the $\boldsymbol{f}_i$ and $\boldsymbol{f}_j$ pairs (i.e., $w_{ij}$ is the $i$th row and $j$th column entry of the $\boldsymbol{W}$ matrix). In other word, when $\boldsymbol{f}_i$ is a representative for $\boldsymbol{f}_j$, the corresponding row in the $\boldsymbol{W}$ matrix is non-zero. Since our goal is to find some few representations of $\boldsymbol{f}_j$ using $\boldsymbol{f}_i$, we impose a row-sparsity constraint on the $\boldsymbol{W}$ matrix in order to select only a few $\boldsymbol{f}_i$s and set the other rows of $\boldsymbol{W}$ entirely equal to zero. 

In order to achieve this goal, we exploit a joint $\l_{p,q}$ norm regularization (the second constraint in Eq. \ref{eq:first main opt eq}),
where $\lambda$ is a regularization parameter and determines the number of non-zero rows of the $\boldsymbol{W}$ matrix. The $\l_{pq}$ norm, is convex for $p\geq1$ and $q\geq1$; otherwise it is a quasi-norm
and is non-convex. 
In fact, we can consider any $q\geq1$, however, $\l_{p,\infty}$ (i.e., $q=\infty , p\leq1$) has the property of giving us the real number of non-zero features which is the desired goal in our feature selection task. 
The $\l_{p,\infty}$  penalty is a convex relaxation of a pseudo-norm which counts the number of non-zero rows in $\boldsymbol{W}$. Another consideration in $\l_{pq}$ norm is the choice of $p$. Some works such as \cite{chartrand2008restricted} have investigated the $\l_p$ norm with $0<p<1$. It is worth noting that for $0<p<1$, Eq. \ref{eq:first main opt eq} is not a convex problem and we cannot guarantee the global minimum and the solution is not unique and it highly depends on initialization. In other words, even though $0<p<1$ might lead to more sparse result, but the solution would not be consistent. Additionally, choosing an optimum initialization method is not straight forward. Hence, in this work, we consider $p=1$ that leads to a convex problem and global minimum for the optimization problem in Eq. \ref{eq:first main opt eq}. As a result, the proposed optimization problem will have the following form:

\begin{equation} \label{eq:second main opt eq}
\centering
\begin{gathered}
\operatornamewithlimits{min}\limits_{w_{ij}}{\sum_{j=1}^{N}\sum_{i=1}^{N}(\sum_{k=1}^{m}(( \boldsymbol{f}_i(k)-\boldsymbol{f}_j(k)) \ln (\frac{\boldsymbol{f}_i(k)}{\boldsymbol{f}_j(k)}))w_{ij}}   \\
s.t.~ \sum_{i=1}^{N} w_{ij}=1,~\forall{j}; ~~
\sum_{i=1}^{N} (\operatornamewithlimits{max}\limits_{1\leq j\leq N}{w_{ij}}) \leq \lambda \\~~~w_{ij}\geq{0}, ~\forall{i,j}, 
%~~~\Arrowvert{W}\Arrowvert_{p,q}=k
\end{gathered}
\end{equation}

We refer to Eq. \ref{eq:second main opt eq} as the Divergence-based Feature Selection (DFS) method. 
We select the feature histograms, corresponding to indices of non-zero rows of $\boldsymbol{W}$ as our representative features in each class and we repeat this process for all the classes in the training set. The number of selected features for each class is determined by the regularization parameter $\lambda$ (i.e., $\lambda$ is roughly the number of non-zero rows in the $\boldsymbol{W}$ matrix). It is important to note that the value of $\lambda$ should satisfy $\lambda \leq N$ (i.e., $N$ is the number of training data in each class), otherwise the $\boldsymbol{W}$ matrix would be the identity matrix, since each probability distribution $\boldsymbol{f}_i$ is the best representation for itself. The convex optimization problem in Eq. \ref{eq:second main opt eq} is solved using the Alternating Direction Method of Multipliers framework in \cite{boyd2011distributed}.

\subsection{Generating Low Dimensional Feature Codes} \label{SubSec: Encoding}

After constructing a compact representation of the feature histograms for all the classes in the training set, it will be saved in the smart cameras' memory (we refer to this compact representation as $\boldsymbol{D}$). In the on-line testing stage in each camera, a feature histogram $\boldsymbol{h}_i$ is extracted for the $i$th test image at each of the $p$ local cameras independently, and we calculate the corresponding code (i.e., $\boldsymbol{s}_i$) for each feature histogram, using a supervised constrained non-negative matrix factorization scheme:  

\begin{equation}
\begin{gathered}
\operatornamewithlimits{min}\limits_{\boldsymbol{s}_i}\{\left\|\boldsymbol{h}_i - \boldsymbol{D}\boldsymbol{s}_i\right\|^2_2 \},~~ i=(1,...,p),\\
~s.t.~ ~\boldsymbol{s}_i(j) \geq {0},~ j=1, \dots, k, ~~~~~~ \sum_{j=1}^{k} \boldsymbol{s}_{i}(j)=1
\label{eq:NNLS1}
\end{gathered}
\end{equation}
where $ \boldsymbol{h}_i\in R^{m\times 1}$, $ \boldsymbol{D}\in R^{m\times k}$, $ \boldsymbol{s}_i\in R^{k\times 1}$ and $ k<<m $.
$k$ is the number of features that have been selected for the whole training set in the previous step (i.e., number of columns of $\boldsymbol{D}$), and $m$ is the dimension of the original feature histograms (i.e., $1000$ in our setting).
The optimization problem in Eq. \ref{eq:NNLS1} is developed from the nonnegative constrained least squares (NCLS) method \cite{chang2000constrained} in conjunction with the sum-to-one constraint. The objective is to minimize the least squares error:
\begin{equation}
\begin{gathered}
\operatornamewithlimits{min}\limits_{\boldsymbol{s}_i}\left\|\hat{\boldsymbol{h}_i }- \hat{\boldsymbol{D}}\boldsymbol{s}_i\right\|^2_2 ,~~ i=(1,...,p),\\
s.t.~ ~\boldsymbol{s}_i(j) \geq {0},~ j=1, \dots, k, 
\label{eq:NNLS2}
%\end{aligned}
\end{gathered}
\end{equation}
where $\hat{\boldsymbol{h}}_i$ and $\hat{\boldsymbol{D}}$ are the augmented matrices
\begin{equation}
\small{
\begin{aligned}
\hat{\boldsymbol{h}}_i=\left[\begin{array}{c}\delta \boldsymbol{h}_i\\ \boldsymbol{1}\end{array}\right],~~ \hat{\boldsymbol{D}}=\left[\begin{array}{c}\delta \boldsymbol{D}\\ \boldsymbol{1}^T\end{array}\right]
\label{eq:NNLS3}
\end{aligned}}
\end{equation}
with $\delta$ being a small weight and $\boldsymbol{1}^T$ is a row vector of all 1’s. This augmentation is used to incorporate the sum-to-one constraint. The constrained minimization problem in Eq. \ref{eq:NNLS2} is solved by a standard active set method \cite{bro1997fast}. This process is simple and can be done fast inside each smart camera. 
After finding $\boldsymbol{s}_i \in R^ {k\times 1}$ in each camera ($i=1,\dots, p$), these low dimensional codes will be sent to the base station for performing the intended recognition task. Transmitting codes with $k$ dimension instead of feature histograms with $m$ dimension leads to major saving in bandwidth of the wireless network (the compression ratio: $m/k,~ k<<m$) as well as better recognition accuracy.   
\begin{table*}[h!]
	\label{table}
	\centering
	\caption{\small Recognition accuracy comparison on BMW dataset. Compression ratio: $2.4$}
	\tiny {
	\begin{tabular}{|c|c|c|c|c|c|c|c|c|c|c|c|c|c|c|c|c|c|c|c|c|c|}
		\hline 
		Class &1 &2  &3  &4  &5  &6  &7  &8  &9  &10  &11  &12  &13  &14  &15  &16  &17  &18  &19  &20&Avg.  \\ 
		\hline 
		DFS & \textbf{99} & 86 & \textbf{86} & \textbf{100} & 86 & \textbf{93} & \textbf{98} & \textbf{94} & \textbf{91} & \textbf{76} & \textbf{96} & \textbf{99} & \textbf{86} & \textbf{100} & \textbf{86} & \textbf{100} & \textbf{100} & \textbf{100} & \textbf{100} & 93&\textbf{93.55} \\ 
		\hline 
		SPCA & 94.44& \textbf{91.66} & 66.66 &81.94  &\textbf{91.66}  &88.88  &93.05  &91.66  &73.61  &65.27  &76.38  &83.33  & 72.22 &93.05  &80.55  &79.16  &90.27  &93.05  &83.33  &100& 84.51 \\ 
		\hline 
		SfM & 83.33&90.27  &58.33  &65.27  &81.94  &87.50  &86.11  &72.22  &63.88  &61.11  &69.44  &70.83  &52.77  &90.27  &75.00  &80.55  &84.72  &\textbf{100}  &86.11  &\textbf{95.83}  &77.77\\ 
		\hline 
	\end{tabular} }
\end{table*}
\section{Experiments and Results} \label{Sec: Exp}
In this work, we validate our proposed feature encoding scheme on two multi-view recognition tasks including pedestrian recognition in surveillance video and distributed object recognition in smart camera networks. 
\subsection{Datasets}
\begin{figure}[h]
\centering
\includegraphics[width=0.4\linewidth]{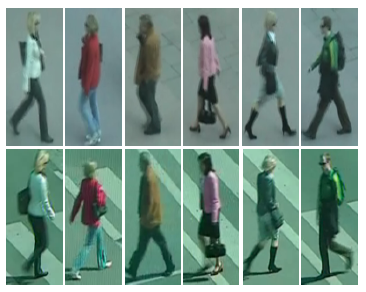}
\caption{\small PRID dataset samples}
\label{fig:priddataset}
\end{figure}

The Person Re-ID (PRID) dataset \cite{PREID_data} is one of the few multi-view datasets which includes multi image frames for each pedestrian recorded from two different, static surveillance cameras. Images from these cameras contain a viewpoint change and a stark difference in illumination, background and camera characteristics. Since images are extracted from trajectories, several different poses per pedestrian are available in each camera view. It contains recorded frames of 475 person trajectories from one view and 856 from the other one, with 245 persons appearing in both views \cite{PREID_data}. Figure \ref{fig:priddataset} illustrates some sample frames of this dataset. 
The second dataset is the Berkeley Multi-view Wireless (\textit{BMW}) database \cite{naikal2010towards} that consists of multiple-view images of $ 20 $ landmark buildings on the Berkeley campus. We employ this dataset in order to evaluate the performance of our proposed algorithm on multi-view object recognition. It is important to note that the image quality in this database is considerably lower than many existing high-resolution databases, which is intended to reproduce realistic imaging conditions for surveillance applications \cite{naikal2010towards}.
%\subsection{Distributed Recognition Experiment}
%In our experiments, dense SIFT feature descriptors are computed at a grid of overlapped image patches. These invariant features are further quantized to form a dictionary of visual words using bag-of-words (\textit{BoW}) approach \cite{lee2008libpmk}. Using hierarchical k-means, all the feature descriptors are clustered into visual words. Then a weighted visual histogram is defined for each image \cite{V-TREE-nister2006scalable}. Each image histogram is a $ 1000 $-D vector and is calculated for all the training and testing images. 
%The feature histogram of the test object is sent to the base station and a nearest neighbor search is conducted in the training set to find the closest histogram for the testing sample. The class label of the closest histogram in the training set is then assigned  to the corresponding test data. 
%In order to fasten the search procedure, we cluster the data into a hierarchical of clusters and do a depth-first search to find the approximate nearest-neighbor to the query point. The chi-square distance is utilized as closeness measure of two $L1$-normalized histograms in our Nearest Neighbor classifier.

%\section{Results} \label{Sec: Results}
\subsection{Pedestrian Recognition in Surveillance Video}
\begin{figure}[h]
\centering
\includegraphics[width=0.9\linewidth]{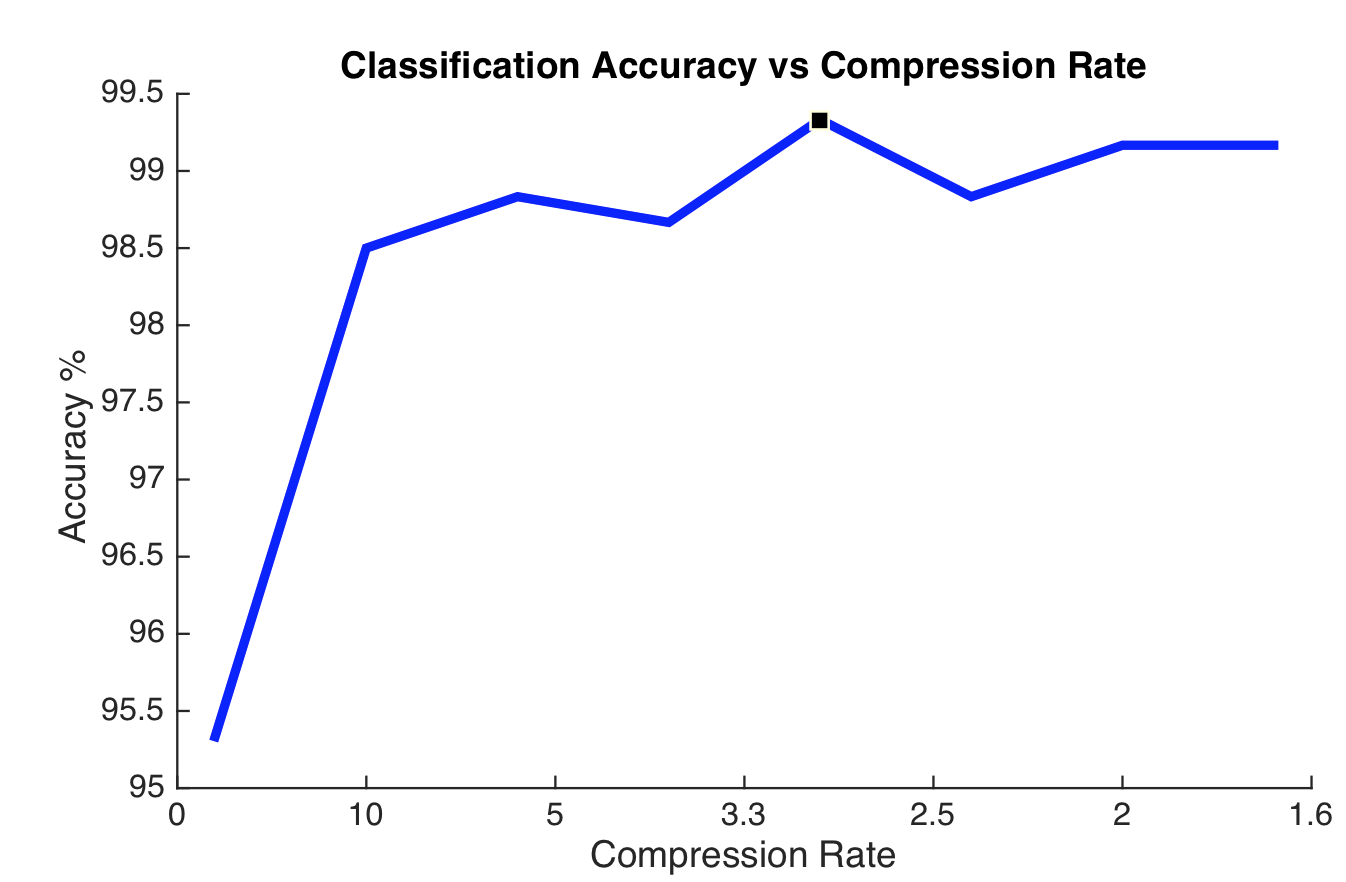}
\caption{\small {Recognition accuracy for different dimensions of the feature histograms using the DFS method}}
\label{fig:accvsdim}
\end{figure}

\begin{figure}
\centering
\includegraphics[width=0.8\linewidth]{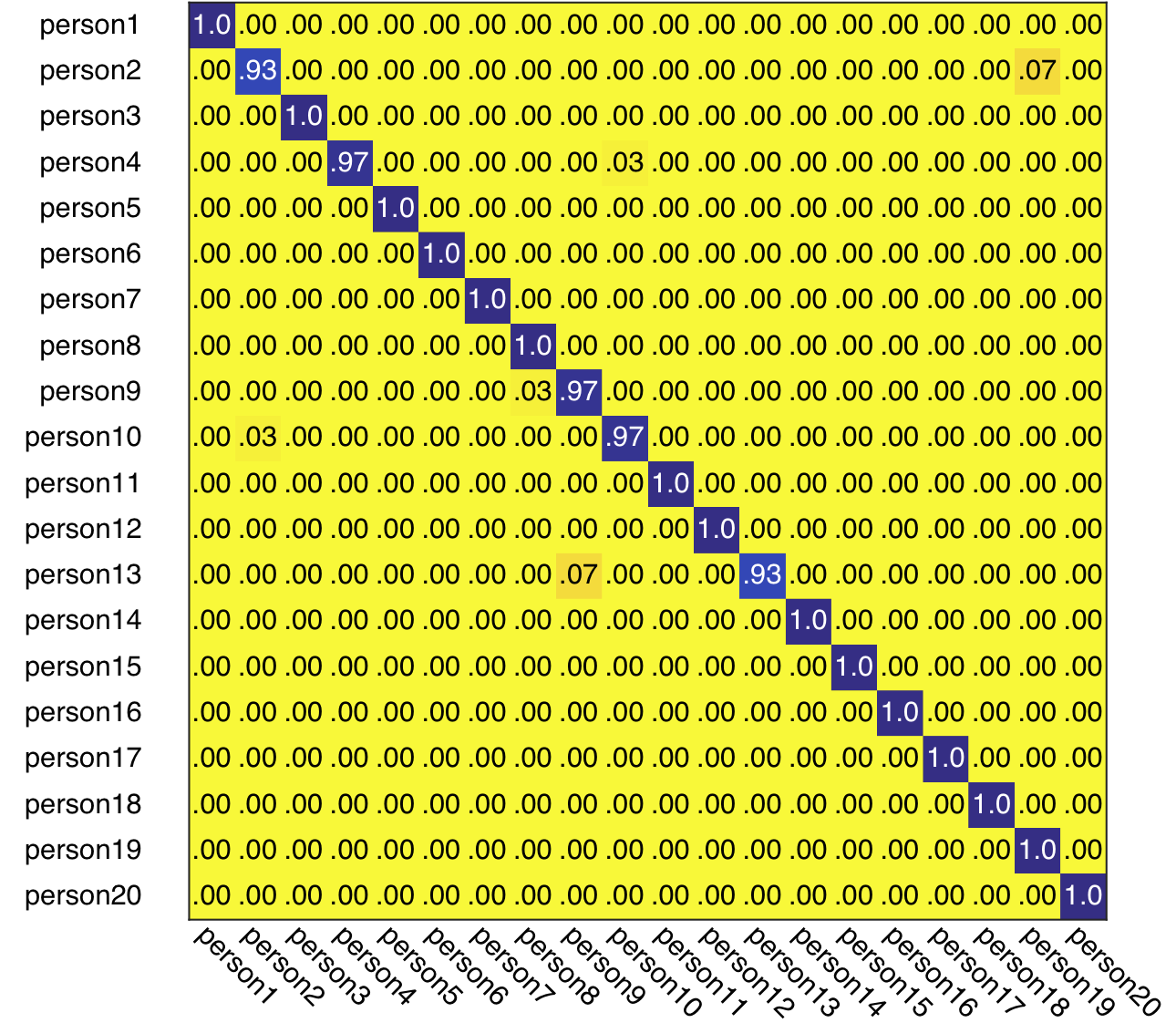}
\caption{\small {Confusion matrix of pedestrian recognition via DFS on PRID dataset using $340$-D features.}}
\label{fig:confm}
\end{figure}

In the first experiment on PRID dataset, we consider $30$ different frames for each pedestrian in each camera and we randomly choose $20$ persons in the dataset for the recognition task. Hence, there are $1200$ images for which we randomly pick half of it as the training set and the rest as the testing set.
The dimension of the original feature histograms is $1000$. Figure \ref{fig:accvsdim} shows the recognition accuracy versus the compression rate using the proposed DFS method. 
\vspace{0 cm}
In this figure, we can observe that with compression rate of 2.94 (i.e., features with dimension of $340$), the accuracy is slightly better than using the original features. The reason is that our feature selection scheme omits those features which are closer to the features from other classes than the features in their own class. We define the compression ratio as dimension of the original features divided by dimension of the encoded features (i.e., $k$ in Eq. \ref{eq:NNLS1}). For instance, in Figure \ref{fig:accvsdim} at the marked point on the curve with compression ratio of $2.94$, the feature dimension is equal to $1000/2.94=340$. Figure \ref{fig:confm} illustrates the recognition accuracy for classification of all the $20$ persons in the experiment with $340$-D feature histograms ($\lambda=17$ in Eq. \ref{eq:second main opt eq}). 
It is worth noting that the recognition task in this experiment is different from person re-identification task which is based on image matching and retrieval.   
\subsection{Building Recognition in BMW Dataset}
In the second experiment (on BMW), there exist $ 16 $ different vantage points, and at each vantage point, images are taken by five cameras simultaneously, thereby summing to $ 80 $ images per category.\vspace{0 cm}
In this section, we compare the recognition accuracy of DFS method with two other existing works. Table 1 demonstrates the classification accuracy of different methods based on Sparse PCA (SPCA) \cite{naikal2011informative} and Structure from Motion (SfM) \cite{turcot2009better}. To have a fair comparison, we set up the same experimental environment as the other two works. In fact, we only considered 8 images (even vantage points of camera $\#2$) from each object for training and the rest of images from other cameras for testing (and compression ratio: 2.4). For most of the object categories our proposed method, outperforms SPCA and SfM based approaches.
One important reason for outperforming the proposed DFS method compared to other two methods is that in contrast to SPCA and SfM methods, the physical
interpretation of the reduced space is preserved during the dimensionality reduction procedure which is critical in the recognition task.  
\section{Conclusion} \label{Sec: Conclusion}
In this paper, we introduced a probabilistic encoding approach based on divergence of the probability distributions of the visual features in limited bandwidth distributed camera networks. The performance of the proposed approach was discussed in two surveillance recognition tasks. The proposed DFS approach is applicable to the variety of distributed computer vision tasks based on transmission of the visual features in a network (e.g., cross view action recognition, person re-identification, etc.). 
% -------------------------------------------------------------------------
\small
\bibliographystyle{IEEEbib}
\bibliography{refs}

\end{document}